\documentclass[conference]{IEEEtran}
\IEEEoverridecommandlockouts
\usepackage{cite}
\usepackage{amsmath,amssymb,amsfonts}
\usepackage{algorithmic}
\usepackage{graphicx}
\usepackage{makecell}
\usepackage{textcomp}
\usepackage{xcolor}
\usepackage{url} 
\usepackage{multirow}
\usepackage{adjustbox}
\usepackage{booktabs}
\def\BibTeX{{\rm B\kern-.05em{\sc i\kern-.025em b}\kern-.08em
    T\kern-.1667em\lower.7ex\hbox{E}\kern-.125emX}}

\begin{document}

\title{Understanding Fairness of Gender Classification Algorithms Across Gender-Race Groups\\
}

\author{\IEEEauthorblockN{Anoop Krishnan, Ali Almadan, Ajita Rattani}
\IEEEauthorblockA{\textit{Dept. of Electrical Eng. and Computer Science} \\
\textit{Wichita State University
Wichita, USA}\\
axupendrannair@shockers.wichita.edu;~ajita.rattani@wichita.edu}
}

\maketitle

\begin{abstract}
Automated gender classification has important applications in many domains, such as demographic research, law enforcement, online advertising, as well as human-computer interaction. Recent research has questioned the fairness of this technology across gender and race.  Specifically, the majority of the studies raised the concern of higher error rates of the face-based gender classification system for darker-skinned people like African-American and for women.
However, to date, the majority of existing studies were limited to African-American and Caucasian only. The aim of this paper is to investigate the differential performance of the gender classification algorithms across gender-race groups. To this aim, we investigate the impact of (a) architectural differences in the deep learning algorithms and (b) training set imbalance, as a potential source of bias causing differential performance across gender and race. Experimental investigations are conducted on two latest large-scale publicly available facial attribute datasets, namely, UTKFace and FairFace. The experimental results suggested that the algorithms with architectural differences varied in performance with consistency towards specific gender-race groups. For instance, for all the algorithms used, Black females (Black race in general) always obtained the least accuracy rates. Middle Eastern males and Latino females obtained higher accuracy rates most of the time. Training set imbalance further widens the gap in the unequal accuracy rates across all gender-race groups. Further investigations using facial landmarks suggested that facial morphological differences due to the bone structure influenced by genetic and environmental factors could be the cause of the least performance of Black females and Black race, in general.


\end{abstract}


\section{Introduction}
Automated facial analysis (FA) includes a wide range of applications, including face detection~\cite{dlib}, visual attribute classification such as gender and age prediction~\cite{Levi15}, and actual face recognition~\cite{8953996}.
 
Among other visual attributes, \emph{gender} is an important demographic attribute~\cite{Ricanek06,Levi15}.
Gender classification refers to the process of assigning male and female labels to biometric samples. Automated gender classification has drawn significant interest in numerous applications such as surveillance, human-computer interaction, anonymous customized advertisement system, and image retrieval system. In the context of biometrics, gender can be viewed as a soft biometric trait~\cite{jain04}  that  can  be  used to  index databases  or  to  enhance  the  recognition  accuracy  of  primary biometric traits such as face and ocular region. Companies such as IBM, Amazon, Microsoft, and many others have released commercial software containing automated gender classification system. 
\begin{figure}[!t] 
    \centering
    \includegraphics[width=0.4\textwidth]{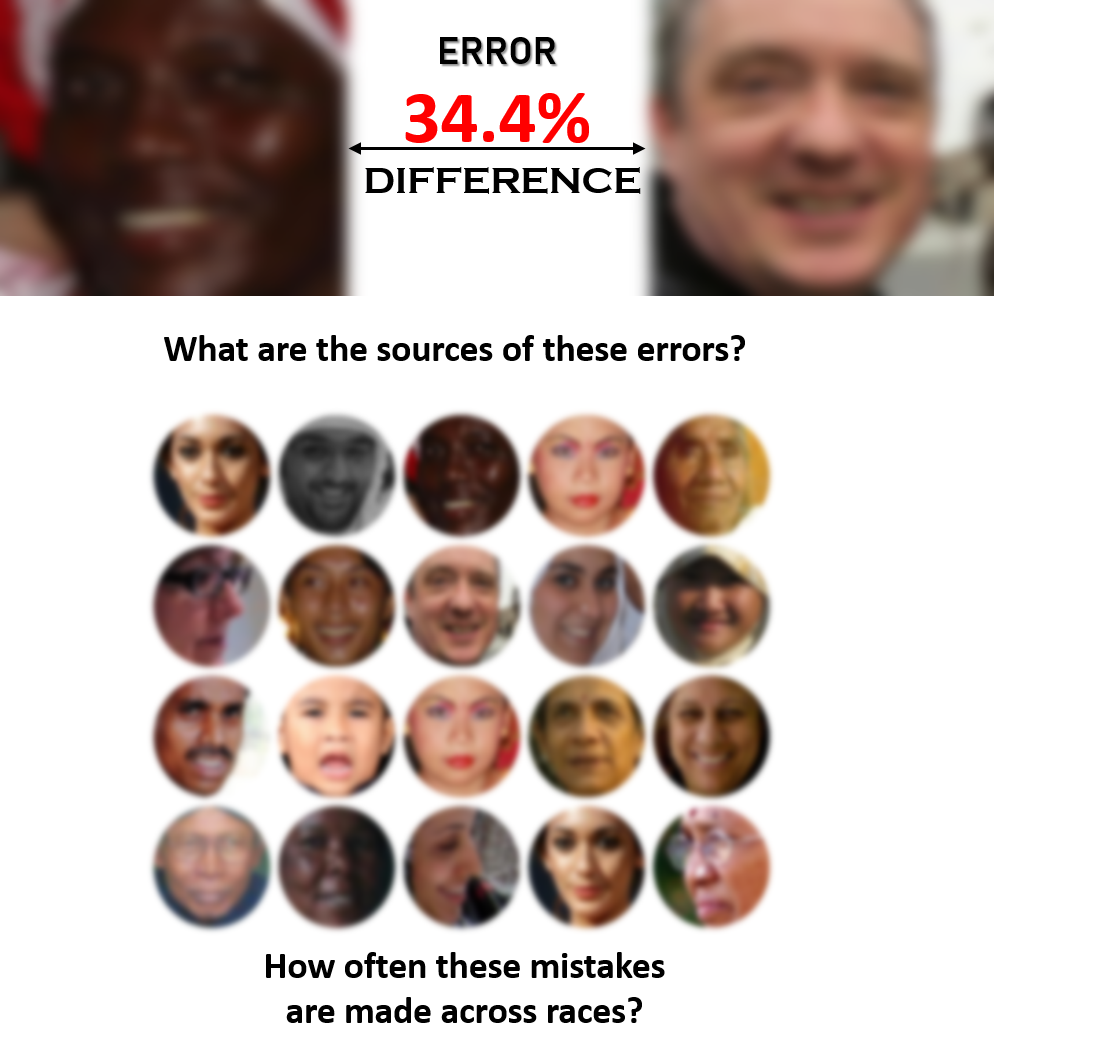}
    \caption{This figure highlights real problems in gender classification algorithms produced by commercial vendors. However, such mistakes have not been studied in a systematic way to understand their underlying causes.}
    \label{SchemeFairness}
\end{figure}

According to ISO/IEC 22116~\cite{ISO_gender}, the term “gender” is defined as “the state of being male or female as it relates to social, cultural or behavioural factors”, the term “sex” is understood as “the state of being male or female” as it relates to biological factors such as DNA, anatomy, and physiology.  Therefore, the term “sex” would be more appropriate instead of “gender” in the context of this study. However, in consistency with the existing studies~\cite{Levi15,Ricanek06,smith2018transfer,Buolamwini18}, the term gender is used in this paper instead of sex.

Over the last few years, the fairness of the gender classification system has been questioned~\cite{Buolamwini18,Muthukumar19,krkkinen2019fairface}. Fairness is the absence of any prejudice or favoritism toward an individual or a group based on their inherent or acquired characteristics~\cite{mehrabi2019survey}. Thus, an unfair (biased) algorithm is one whose decisions are skewed towards a particular group of people. The problem of unequal accuracy rates has been highlighted in gender classification from face images for dark-skinned people and women~\cite{Buolamwini18,Muthukumar19}. 

Specifically, a research study by the MIT Media Lab~\cite{Buolamwini18} uncovered substantial accuracy differences in face-based gender classification tools from companies like Microsoft, IBM, Face++, and Amazon~\cite{AI, FA}, with the lowest accuracy for \textbf{dark-skinned females}. The underlying cause of the unequal misclassification rates in gender classification is not investigated in this study. Muthukumar~\cite{Muthukumar19} analyzed the influence of the skin type on gender classification accuracy and concluded that the skin type has a minimal effect on classification decisions. However, the dataset used~\cite{Muthukumar19} consisted only of African-American and Caucasian.

Some of the \emph{limitations} of the published research~\cite{Buolamwini18,Muthukumar19} in relation to the fairness of the face-based gender classification are as follows:
\begin{itemize}
    \item \textbf{Limited investigation}: There is a lack of understanding of the cause(s) of demographic variation in the accuracy of the gender classification system. 
    \item \textbf{Limited dataset evaluation}: Mostly limited size datasets consisting of a limited number of races, mostly African-American and Caucasian, are used for evaluation. 
    \item \textbf{Black-box evaluation}: Commercial SDKs from IBM, Face++, and Amazon are used for the fairness evaluation of face-based gender classification system. Therefore, sources of bias may not be ascertained.
\end{itemize}

It is still not clear how the error propagates across multiple gender-race groups for different gender classification algorithms. It is also unknown if the errors are due to skewed training dataset or algorithmic bias (caused by the inherent structure of the algorithm).  Figure~\ref{SchemeFairness} highlights the problems in current gender classification algorithms.

With the widespread use of gender classification system, it is essential to consider fairness issues while designing and engineering this system.  The fairness is a compelling social justice as well as an engineering issue. In order to address the bias issue in the gender classification system, it is important to investigate its \emph{source}.

\subsection{Our Contribution}
In order to further improve understanding of the fairness of the face-based gender classification system across races. Our contributions are the following:

\begin{itemize}
    \item \textbf{Investigating the sources of bias}: The impact of training set imbalance and architectural differences in algorithms are analyzed. Further, the facial morphological differences obtained using $68$ facial landmark coordinates~\cite{dlib} are analyzed in understanding the cause of differential accuracy for specific gender-race groups (i.e., Black females).
    \item \textbf{Thorough evaluation on large-scale datasets}: All the analyses are conducted on the latest UTKFace~\cite{zhifei2017cvpr} and FairFace~\cite{krkkinen2019fairface} facial attributes datasets consisting of four and seven race groups, respectively. Apart from accuracy values, false positives and false negatives are also analyzed.
    \item \textbf{White-box evaluation}: Open-source deep learning based gender classification algorithms are evaluated for full access to algorithms and training data.

    
\end{itemize}

This paper is organized as follows: Section II discusses the prior work in deep learning-based algorithms for gender classification and the study on its fairness analysis. Section III discuss the CNNs used in this study for gender classification. Experimental evaluations and the obtained results are discussed in section IV. Conclusion and future work are discussed in section V.

\section{Prior Work}
This section discusses the recent literature on deep learning-based gender classification from facial images and the related study on its fairness analysis.

\subsection{CNNs for Gender Classification from Facial Images}
A Convolution Neural Network (CNN) is a type of feed-forward artificial neural network in which the connectivity pattern between its neurons, that have learnable weights and biases, is inspired by the organization of the visual cortex. The efficacy of CNNs has been very successfully demonstrated for large scale image recognition~\cite{ILSVRC15}, pose estimation, face recognition, and face-based gender classification~\cite{Levi15}, to name a few.

In~\cite{Levi15}, an end-to-end CNN model was evaluated on the Adience benchmark. The average gender classification accuracy of $88.1\%$ was reported. Further, studies used fine-tuned~\cite{8265401} VGG, InceptionNet, and ResNet (pretrained on ImageNet dataset~\cite{ILSVRC15}) for gender classification from facial images. Specifically, pretrained ImageNet models are fine-tuned on the datasets annotated with gender labels. Gender classification accuracy in the range [$87.4\%$, $92.6\%$] was obtained on Adience dataset. The authors concluded that different CNN architectures obtained different results. Fine-tuned CNNs obtained better results over those trained from scratch. In~\cite{smith2018transfer}, transfer learning was explored using both VGG-19 and VGGFace for gender classification on the MORPH-II dataset. Accuracy of $96.6\%$ and $98.56\%$ was obtained for VGG19 and VGGFace, respectively. The higher performance of VGGFace was attributed to pretrained weights obtained from facial images.

In~\cite{DBLP:journals/corr/LiuLWT14}, authors proposed a novel deep learning framework for attribute prediction in the wild. It cascades two CNNs, LNet and ANet, which are fine-tuned jointly with attribute tags, but pre-trained differently. LNet is pre-trained by massive general object categories for face localization, while ANet is pre-trained by massive face identities for attribute prediction. The maximum of $94\%$ accuracy was obtained on CelebA dataset.

The above-mentioned studies evaluated the overall accuracy. The fairness of the gender classification model across males and females was not evaluated. In fact, the datasets such as Adience~\cite{Levi15,8265401} and CelebA~\cite{DBLP:journals/corr/LiuLWT14} often used in the existing studies revealed over-representation of lighter and under-representation  of  darker  individuals in general. For instance, $86.2\%$  of  the  subjects in the Adience benchmark~\cite{Levi15} consists of lighter-skinned individuals.

\subsection{Fairness of the Gender Classification System}
Buolamwini and Gebru~\cite{Buolamwini18} evaluated fairness of the gender classification system using three commercial SDKs from Microsoft, Face++, and IBM on Pilot Parliaments Benchmark (PPB) developed by the authors. The dataset consists of $1270$ individuals from Africans and European races, and the female and male contribution was $44.6\%$ and $55.4\%$, respectively. The accuracy differences of $23.8\%$, $36.0\%$, and $33.1\%$ was obtained for dark-skinned females using Microsoft, Face++, and IBM, respectively.

Muthukumar~\cite{Muthukumar19} analyzed the influence of the skin type for understanding the reasons for unequal gender classification accuracy on face images. The skin type of the face images in the PPB dataset was varied via color-theoretic methods, namely luminance mode-shift and optimal transport, keeping all other features fixed. The open-source convolutional neural network gender classifier was used for this study. The author concluded that the effect of skin type on classification outcome is minimal. Thus, the unequal accuracy rates observed in~\cite{Buolamwini18} is likely not because of the skin type. \emph{However, only African American and Caucasian are used in this study}.

Worth-mentioning that both the above studies~\cite{Buolamwini18,Muthukumar19} used the PPB dataset consisting of  $1270$ subjects from Africans and Europeans. Studies in~\cite{Smith_2020_WACV, Ryu18,serna2020insidebias} also proposed data augmentation, two-fold transfer learning and measuring bias in deep representation to mitigate its impact in biometric attribute classifier (such as gender and age). In an attempt to advance the state-of-the-art in the fairness of facial analysis methods, face attribute dataset for the balanced race, gender, and age classification was assembled in 2019~\cite{krkkinen2019fairface}. The authors showed the performance of the ResNet model trained on this dataset for gender, age, and race classification. The average accuracy of $94.4\%$ was obtained on the gender classification model when tested on an external testbed.








\section{Convolutional Neural Network (CNN) Models Used}
\begin{figure}
    \centering
    \includegraphics[width=0.33\textwidth]{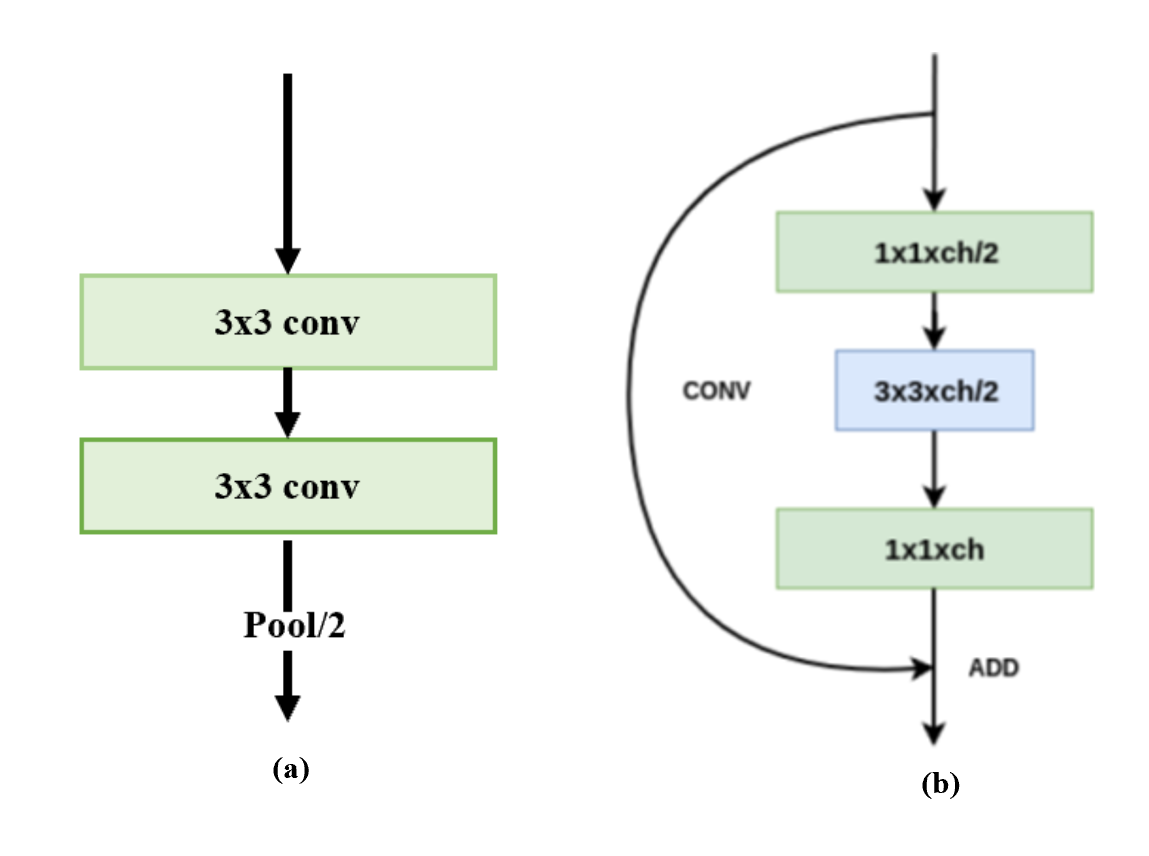}
    \includegraphics[width=0.33\textwidth]{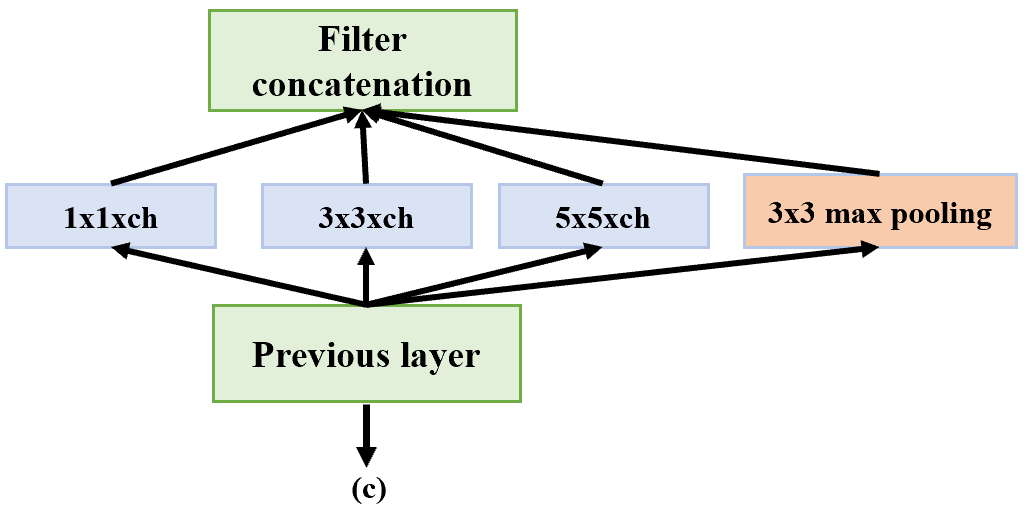}
             \caption{Architectures of the (a) VGG, (b) ResNet, and (c) InceptionNet models used in this study.}
    \label{fig:arc}
\end{figure}

This section discuss the deep-learning based CNN models fine-tuned for gender classification. These CNN models are pre-trained on large scale ImageNet~\cite{ILSVRC15} dataset comprising of $1.2$ million training images and have become the standard
benchmark for large-scale image classification. Figure~\ref{fig:arc} shows architecture of these CNN models.

\begin{enumerate}
\item \textbf{VGG}: The VGG architecture was introduced by Visual Graphics Group~(VGG) research team at Oxford University~\cite{Simonyan14c}. The architecture consists of sequentially stacked $3\times 3$ convolutional layers with intermediate max-pooling layers followed by a couple of fully connected layers for feature extraction. Usually, VGG models have $13$ to $19$ layers. We used VGG-16 and VGG-19 in this study which has $138M$ and $140M$ number of parameters. We also evaluated VGGFace model which is basically VGG-16 trained on VGGFace2 dataset~\cite{8373813}.


\item \textbf{ResNet}: ResNet is a short form of residual network based on the idea of  ”identity  shortcut  connection”  where input  features  may  skip  certain layers~\cite{he2016deep}.  In this study, we used ResNet-$50$ which has $23.5$M parameters.

\item \textbf{InceptionNet}: The hallmark of this network~\cite{7298594} is its carefully crafted design: the depth and width of the network is increased while keeping the computational requirements constant. The architecture has a total of
$9$ Inception modules, which allow for pooling and convolution operation with different filter sizes to be performed in parallel. In this study, we used InceptionNet-v4. 
\end{enumerate}

\textbf{Network Implementation and Fine-tuning:}
We used pytorch (https://pytorch.org/) implementation of these pretrained networks (VGG-16, VGG-19, VGGFace, ResNet-50 and InceptionNet-v4) along with their weight files for fine-tuning them. 
These networks were fine-tuned for gender classification using training set of facial images annotated with gender labels (male and female). Fine-tuning was done by extracting all the layers but the last fully connected layers from aforementioned pre-trained networks and adding new fully connected layer(s) along with softmax.  

Based on empirical evidence on validation set, fine tuning of the VGG architectures and ResNet was performed by an additional two 512-way fully connected layers and one 2-way output layer (equal to the number of classes) along with softmax layer. For InceptionNet-v4, all the layers were extracted until
the fully connected layer followed by additional 4096-way, 512-way
and one 2-way output layer along with softmax. 
The fine-tuning was performed using Stochastic Gradient Descent (SGD) optimizer with an initial learning rate of $0.0001$ for $1000$ epochs using early stopping mechanism.



\section{Experimental Evaluation}
In this section, datasets used, experiments conducted and the results obtained are discussed. We used cropped face images obtained using Dlib face detection utility~\cite{dlib}.
\subsection{Datasets}
\begin{figure} [h]
    \centering
    \includegraphics[width=0.4\textwidth]{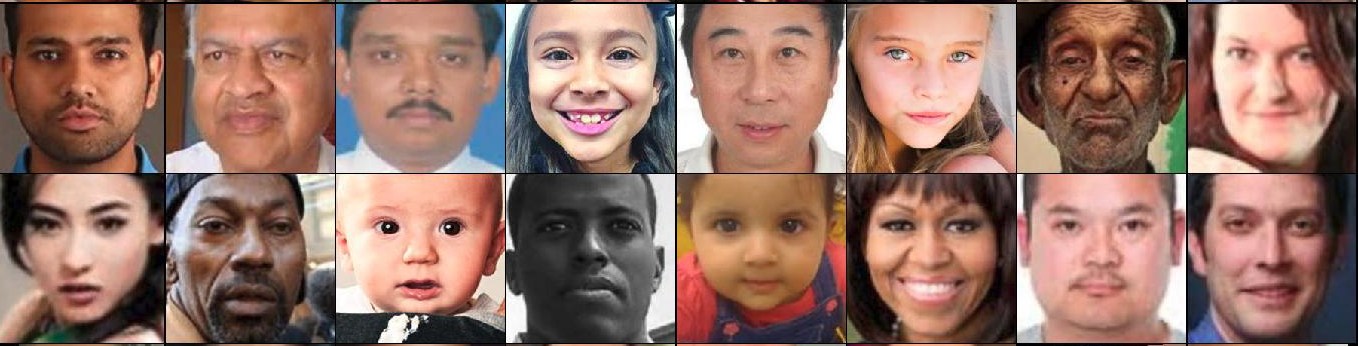}
            \caption{Sample images from UTKFace dataset~\cite{zhifei2017cvpr}.}
    \label{fig:samUTK}
\end{figure}

\textbf{UTKFace~\cite{zhifei2017cvpr}}: UTKFace~\cite{zhifei2017cvpr} is a large-scale face dataset with long age span (range from $0$ to $116$ years old). The dataset consists of total of $20,000$ face images scrapped from the web and annotated with age, gender, and race labels. The images cover large variation in pose, facial expression, illumination, occlusion, and resolution. The four race groups included are as follows: White, Black, Indian, and Asian.
The training portion of the UTKFace dataset consist of $41\%$ females and $59\%$ males, therefore is skewed towards males.
Table~\ref{dataDis_utk} shows the complete sample distribution of training subset of UTKFace dataset used in our experiments. Sample images from UTKFace dataset are shown in Figure~\ref{fig:samUTK}.

\begin{table}[h]
  \centering
  \caption{Training dataset distribution of UTKFace~\cite{zhifei2017cvpr} used in our experiments.}
 \begin{adjustbox}{width=0.9\columnwidth}
 \begin{tabular}{cccc}\hline
\textbf{Race} & \textbf{Female} & \textbf{Male} & \textbf{Total}  \\ \hline
 \textbf{White} & 1117 ($11\%$) & 1903 ($19\%$)  & 3020 ($30\%$)\\
 \textbf{Black} & 1134 ($11\%$) & 1597 ($16\%$) & 2731 ($27\%$)\\
 \textbf{Asian} & 883 ($9\%$) & 1111 ($11\%$) &  1994 ($20\%$)\\
 \textbf{Indian} & 991 ($10\%$) & 1407 ($14\%$) & 2398 ($23\%$)\\ \hline
\textbf{Total} & 4125 ($41\%$) & 6018 ($59\%$) & 10143 ($100\%$)\\ \hline

\end{tabular}
\end{adjustbox}
\label{dataDis_utk}
    \end{table}
%




\begin{figure}
    \centering
    \includegraphics[width=0.4\textwidth]{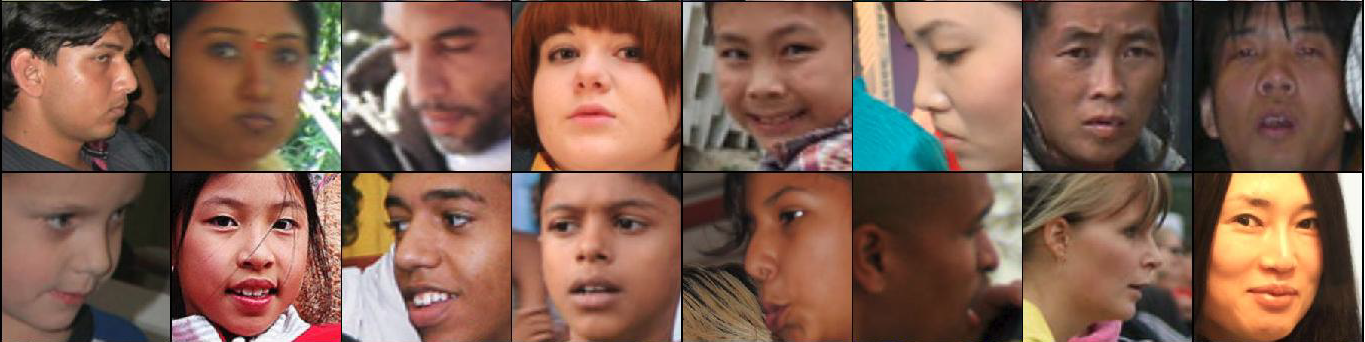}
             \caption{Sample images from FairFace dataset~\cite{krkkinen2019fairface}}
    \label{fig:samFair}
\end{figure}

\textbf{FairFace~\cite{krkkinen2019fairface}}: The facial image dataset consisting of $108,501$ images, with an emphasis on balanced race composition in the dataset~\cite{krkkinen2019fairface}.  The seven race groups defined in the dataset are as follows: White, Black, Indian, East Asian, Southeast Asian, Middle East, and Latino. Images were collected from the YFCC-100M Flickr dataset and labeled with race, gender, and age groups. The dataset was released via~\url{https://github.com/joojs/fairface}. The training portion of the FairFace dataset consist of $47\%$ females and $53\%$ males. Table~\ref{dataDis_fair} shows the complete sample distribution of training$\backslash$test subset of FairFace dataset used in our experiments.
Sample images from FairFace dataset are shown in Figure~\ref{fig:samFair}.


\begin{table}[h!]
  \centering
  \caption{Training$\backslash$Test dataset distribution of FairFace dataset~\cite{krkkinen2019fairface} used in our experiments.}
 \begin{adjustbox}{width=1\columnwidth}
\begin{tabular}{c ccc}\hline
\textbf{Race} & \textbf{Female} & \textbf{Male} & \textbf{Total}\\ \hline
 \textbf{White} & 7826 ($9\%$) $\backslash$ 963 ($9\%$) & 8701 ($10\%$) $\backslash$ 1122 ($10\%$) & 16527 ($19\%$) $\backslash$ 2085 ($19\%$)   \\
 \textbf{Black} & 6137 ($7\%$) $\backslash$ 757 ($7\%$) & 6096 ($7\%$) $\backslash$ 799 ($7\%$) & 12233 ($14\%$) $\backslash$ 1556 ($14\%$) \\
 \textbf{East Asian} & 6141 ($7\%$) $\backslash$ 773 ($7\%$) & 6146 ($7\%$) $\backslash$ 777 ($7\%$) & 12287 ($14\%$) $\backslash$ 1550 ($14\%$)\\
 \textbf{Indian} & 5909 ($7\%$) $\backslash$ 763 ($7\%$) & 6410 ($7\%$) $\backslash$ 753 ($7\%$) & 12319 ($14\%$) $\backslash$ 1516 ($14\%$)\\
 \textbf{ME Eastern} & 2847 ($3\%$) $\backslash$ 396 ($4\%$) & 6369 ($8\%$) $\backslash$ 813 ($7\%$) & 9216 ($11\%$) $\backslash$ 1209 ($11\%$)\\
 \textbf{Latino} & 6715 ($8\%$) $\backslash$ 830 ($8\%$) & 6652 ($8\%$) $\backslash$ 793 ($7\%$) & 13367 ($16\%$) $\backslash$ 1623 ($15\%$)\\
 \textbf{SE Asian} & 5183 ($6\%$) $\backslash$ 680 ($6\%$) & 5612 ($7\%$) $\backslash$ 735 ($7\%$) & 10795 ($13\%$) $\backslash$ 1473 ($13\%$)\\\hline
\textbf{Total} & 40758 ($47\%$) $\backslash$ 5162 ($47\%$) & 45986 ($53\%$) $\backslash$ 5792 ($53\%$) & 86744 ($100\%$) $\backslash$ 10954 ($100\%$)\\ \hline
\end{tabular}
\end{adjustbox}
\label{dataDis_fair}
\end{table}

\subsection{Results}
Following the recommendation in~\cite{8265401,smith2018transfer}, we used fine-tuned models for gender classification.
The deep learning models were fine-tuned on training subset of UTKFace (Table~\ref{dataDis_utk}) and FairFace (Table~\ref{dataDis_fair}) datasets.
$70\%$ of the training data was used for fine-tuning the models, and the rest $30\%$ was used as a validation set. Subjects did not overlap between training and validation sets. 
On average, the training and validation accuracy values of $0.91$ and $0.96$ were obtained for ResNet-50. For VGG architectures, training and validation accuracy values of $0.92$ and $0.97$ were obtained for VGG-19,  $0.93$ and $0.92$ for VGG-16, and $0.98$ and $0.96$ for VGGFace. We also trained AdienceNet model~\cite{Levi15} but due to very low accuracy of $0.65$ on the validation set, this model is not used used for further investigation.
InceptionNet-V4 obtained training and validation accuracy of $0.90$. We even tried training these models from scratch; however, accuracy rates were much lower in comparison to those obtained using fine-tuning, confirming the observation in~\cite{8265401}.
    The fine-tuned models are evaluated on the test subset of the FairFace dataset (Table~\ref{dataDis_fair}) for fairness evaluation across gender-race groups. Next, we discuss the experiments conducted and the results obtained. 

\noindent \textbf{Exp \#1: Training on Gender and Race Balanced Dataset:}
The goal of this experiment is to evaluate and compare the fairness of the different CNN architectures used for gender classification. The hypothesis is different CNN architectures may \emph{obtain different accuracy rates due to feature representation differences emerging owing to their unique architecture}. 
For this experiment, all the CNN models are fine-tuned on gender and race balanced training subset of FairFace (Table~\ref{dataDis_fair}).
The accuracy, false positive, and false negatives of all the models were evaluated and recorded on test subset of FairFace dataset. 

Table~\ref{tab:over_stat_fair} shows the male, female and overall accuracy of the CNN models in gender classification. It can be seen that for most of the models, the overall accuracy of about $91\%$ was obtained on the test set. However, ResNet-50 had a higher male accuracy rate over other models. VGG-16 and VGG-19 obtained higher accuracy rates for females over males. The InceptionNet performed poorly over all other networks. Although InceptionNet obtained training and validation accuracy values of $90\%$, the reason for poor performance on the test set could be over-fitting. 
Among VGGs, VGGFace has higher male accuracy over VGG-16 and VGG-19, this could be attributed to the fact that VGGFace is pretrained on VGGFace2 dataset~\cite{8373813} which is skewed towards male population ($59.7\%$ males and $40.3\%$ females). Therefore, \emph{bias could have been propagated from pretrained weights}. Rest other models used pretrained ImageNet weights obtained from general object classification. There is minimal chance of gender-related bias propagation from ImageNet dataset.

Table~\ref{fairFace:genderrace} shows the accuracy values of all the deep learning models on gender-race groups on FairFace test distribution. It can be seen that despite average accuracy values being equivalent, all the algorithms varied across gender-race groups. 
For instance, ResNet-50 obtained higher accuracy rates for males for all the races. VGG-16 and VGG-19 consistently obtained higher accuracy rates for females for all the races except \emph{Black females} with an average difference of $0.037$ over Black males. VGGFace (which is VGG-16 pretrained on VGGFace2 dataset) obtained higher rates for males except for Latino. InceptionNet-v4 obtained the major difference in the accuracy values between males and females. The least standard deviation of $0.031$ in the accuracy values is obtained by VGG-16 (Table~\ref{fairFace:stats}). In Table~\ref{fairFace:stats}, difference in the average is the mean male and female accuracy values. Overall, \emph{Middle Eastern} males obtained the highest accuracy values followed by Indian and Latino. These results are in accordance with those reported in~\cite{krkkinen2019fairface}. This also suggests that the general notion that White males perform better than others may be incorrect. Latino females obtained the highest accuracy, followed by Middle Eastern females. White and East Asian females obtained equivalent accuracy values overall. \emph{All the models obtained the least accuracy rates for Black females} (average accuracy being $0.749$). 

Further, Table~\ref{fair_FPR_FNR} shows the false positives and false negative of the gender classification system for all the CNN models. False positives are females classified as males, and false negatives are males classified as females.
In accordance with Table~\ref{fairFace:genderrace}, VGG-16, and VGG-19 obtained lower false positives in general, except for Black females. Inception-V4 obtained higher false positives and false negatives. The black race has higher false negatives for most of the models, which means that black females are misclassified as males more often than other females. ResNet-50 maintained a better balance between false positives and false negatives over other models. The highest false negatives are obtained for black, followed by Southeast Asian males meaning that they are more likely to be classified as females. 
         
\emph{Overall, CNN models with architectural differences varied in performance with consistency towards specific gender-race groups. For instance, all the algorithms obtained the least accuracy for Black females and higher accuracy rates for Middle Eastern males. Therefore, the bias of the gender classification system is not due to a particular algorithm} \newline. Study in ~\cite{albiero2020does} also suggest that gender balanced training set did not improve face recognition accuracy for females.



\begin{table}[h]
\centering
\caption{Male, Female and Overall Accuracy of ResNet-50, InceptionNet-V4, VGG-19 and VGGFace for gender classification when fine-tuned on balanced FairFace training dataset.}
\resizebox{\columnwidth}{!}{%
\begin{tabular}{|c|c|c|c|}
\hline
\textbf{Model} & \textbf{Male}  &  \textbf{Female} & \textbf{Overall Accuracy}\\ \hline 
\textbf{ResNet-50}     & 0.94086 &   0.891  &  0.916  \\ \hline 
\textbf{Inception-V4} & 0.871        & 0.559   & 0.7148      \\ \hline 
\textbf{VGG-16}        & 0.916         & 0.923  &  0.919      \\ \hline 
\textbf{VGG-19}        & 0.904        & 0.926  & 0.915    \\ \hline 
\textbf{VGGFACE}      & 0.9245       & 0.908  & 0.916   \\ \hline
\textbf{AVG}    & \textbf{0.911}        &   \textbf{0.841}   & \textbf{0.876}  \\ \hline
\end{tabular}
}
\label{tab:over_stat_fair}
\end{table}




    


\begin{table*}[]
\centering
\caption{Gender classification accuracy of different CNN architectures across gender-race groups. Balanced subset of FairFace dataset was used for fine-tuning the models.}
\begin{adjustbox}{width=1\textwidth}
\begin{tabular}{|l|c|c|c|c|c|c|c|c|c|c|c|c|c|c|}
\hline
\multicolumn{1}{|c|}{\textbf{Race}}   & \multicolumn{2}{c|}{\textbf{WHITE}}                    & \multicolumn{2}{c|}{\textbf{ME EASTERN}}                  & \multicolumn{2}{c|}{\textbf{BLACK}}                       & \multicolumn{2}{c|}{\textbf{EAST ASIAN}}                  & \multicolumn{2}{c|}{\textbf{SE ASIAN}}                  & \multicolumn{2}{c|}{\textbf{INDIAN}}                      & \multicolumn{2}{c|}{\textbf{LATINO}}               \\ \hline
\multicolumn{1}{|c|}{\textbf{Gender}} & \textbf{M}                 & \textbf{F}                & \textbf{M}                  & \textbf{F}                  & \textbf{M}                  & \textbf{F}                  & \textbf{M}                  & \textbf{F}                  & \textbf{M}                 & F                          & M                           & F                           & M                           & F                    \\ \hline
\textbf{Resnet-50}            & 0.940            & 0.899          & 0.966             & 0.914           & 0.887            & 0.803             & 0.938            & 0.897            & 0.946           & 0.885          & 0.952            & 0.912             & 0.957            & 0.924     \\ \hline
\textbf{Inception-V4}        & 0.862            & 0.590           & 0.918            & 0.601            & 0.860              & 0.440              & 0.846            & 0.590             & 0.840            & 0.531           & 0.890            & 0.510             & 0.880             & 0.650     \\ \hline
\textbf{VGG-16}              & 0.909           & 0.928          & 0.954           & 0.960             & 0.895            & 0.838            & 0.912           & 0.938            & 0.895            & 0.922          & 0.927            & 0.923            & 0.917            & 0.951     \\ \hline
\textbf{VGG-19}              & 0.896          & 0.935          & 0.969             & 0.962            & 0.861            & 0.844            & 0.893            & 0.926            & 0.875           & 0.924           & 0.916           & 0.931           & 0.918           & 0.964    \\ \hline
\textbf{VGGFACE}             & 0.928           & 0.908          & 0.957            & 0.937            & 0.895            & 0.824            & 0.922            & 0.925            & 0.918           & 0.903           & 0.926            & 0.911           & 0.926            & 0.946      \\ \hline
\multicolumn{1}{|c|}{\textbf{AVG}}    & \textbf{0.907 } & \textbf{0.852 } & \textbf{0.953 } & \textbf{0.875 } & \textbf{0.879} & \textbf{0.749 } & \textbf{0.902} & \textbf{0.855} & \textbf{0.895} & \textbf{0.833} & \textbf{0.922} & \textbf{0.837 } & \textbf{0.919} & \textbf{0.887} \\ \hline
\end{tabular}%
}

\end{adjustbox}
\label{fairFace:genderrace}
\end{table*}

\begin{table*}[]
\centering
\caption{Statistics of the accuracy values obtained in Table~\ref{fairFace:genderrace}.}
\begin{adjustbox}{width=1\columnwidth}

\begin{tabular}{|l|c|c|c|c|c|}
\hline
                      & \multicolumn{1}{l|}{\textbf{MAX}}& \multicolumn{1}{l|}{\textbf{MIN}} & \multicolumn{1}{l|}{\textbf{AVG}} & \multicolumn{1}{l|}{\textbf{STDEV}} & \multicolumn{1}{l|}{\textbf{Diff. in. AVG}} \\ \hline
\textbf{ResNet50}     & 0.966                    & 0.803                    & 0.916                    & 0.042                      & 0.351                              \\ \hline
\textbf{Inception-V4} & 0.9176                   & 0.44                     & 0.7148                   & 0.17                       & 2.18                               \\ \hline
\textbf{VGG-16}       & 0.96                     & 0.838                    & 0.919                    & 0.031                      & -0.049                             \\ \hline
\textbf{VGG-19}       & 0.97                     & 0.844                    & 0.915                    & 0.038                      & -0.156                             \\ \hline
\textbf{VGGFACE}      & 0.957                    & 0.824                    & 0.916                    & 0.031                      & 0.1175                             \\ \hline
\end{tabular}%

\end{adjustbox}
\label{fairFace:stats}
\end{table*}

\begin{table*}[]
\centering
\caption{False positives and negatives across gender-race groups for accuracy values obtained in Table~\ref{fairFace:genderrace}. False positives are females misclassified as males and false negatives are males misclassified as females.}
\begin{adjustbox}{width=1\textwidth}

\begin{tabular}{|l|c|c|c|c|c|c|c|c|c|c|c|c|c|c|}
\hline
\multicolumn{1}{|r|}{\textbf{Race}} & \multicolumn{2}{c|}{\textbf{WHITE}}                    & \multicolumn{2}{c|}{\textbf{ME EASTERN}}               & \multicolumn{2}{c|}{\textbf{BLACK}}                   & \multicolumn{2}{c|}{\textbf{EAST ASIAN}}                & \multicolumn{2}{c|}{\textbf{SE ASIAN}}                 & \multicolumn{2}{c|}{\textbf{INDIAN}}                    & \multicolumn{2}{c|}{\textbf{LATINO}}                   \\ \hline
                                    & \textbf{FP}              & \textbf{FN}               & \textbf{FP}              & \textbf{FN}               & \textbf{FP}              & \textbf{FN}              & \textbf{FP}               & \textbf{FN}               & \textbf{FP}               & \textbf{FN}              & \textbf{FP}               & \textbf{FN}               & \textbf{FP}               & \textbf{FP}              \\ \hline
\textbf{ResNet-50}                   & 0.100         & 0.059          & 0.085        & 0.034           & 0.196          & 0.113         & 0.104     & 0.062          & 0.115           & 0.054        & 0.087          & 0.047          & 0.075          & 0.042         \\ \hline
\textbf{Inception-V4}               & 0.410           & 0.140            & 0.400            & 0.082           & 0.560           & 0.140           & 0.411           & 0.154           & 0.470            & 0.160           & 0.500            & 0.105          & 0.346           & 0.120           \\ \hline
\textbf{VGG-16}                      & 0.072         & 0.091          & 0.040         & 0.045         & 0.163         & 0.105         & 0.062         & 0.087        & 0.078          & 0.104         & 0.077       & 0.073        & 0.049           & 0.083     \\ \hline
\textbf{VGG-19}                      & 0.065         & 0.104          & 0.038         & 0.030          & 0.156          & 0.140           & 0.074           & 0.107           & 0.076          & 0.125         & 0.069          & 0.084         & 0.036           & 0.082          \\ \hline
\textbf{VGGFACE}                    & 0.092         & 0.072          & 0.063        & 0.043          & 0.176         & 0.105         & 0.075           & 0.078           & 0.097          & 0.081         & 0.089        & 0.074          & 0.054           & 0.074         \\ \hline
\multicolumn{1}{|c|}{\textbf{AVG}}           & \textbf{0.148 } & \textbf{0.093} & \textbf{0.125} & \textbf{0.047 } & \textbf{0.25 } & \textbf{0.120 } & \textbf{0.145} & \textbf{0.098} & \textbf{0.167} & \textbf{0.105} & \textbf{0.165} & \textbf{0.077} & \textbf{0.112} & \textbf{0.080} \\ \hline
\end{tabular}%
\end{adjustbox}
\label{fair_FPR_FNR}
\end{table*}

\noindent \textbf{Exp \#2: Training on Un-balanced (Skewed) Dataset:}

\begin{table}
\centering
\caption{Male, Female and Overall Accuracy of ResNet-50, InceptionNet-V4, VGG-16 and VGGFace for gender classification when fine-tuned on Skewed UTKFace training dataset.}
\resizebox{\columnwidth}{!}{%
\begin{tabular}{|l|c|c|c|}
\hline
\textbf{Model}                 & \textbf{Male}                          & \textbf{Female}                        & \textbf{Overall Accuracy}                      \\ \hline
\textbf{ResNet-50}         & 0.885                      & 0.693                      & 0.789                 \\ \hline
\textbf{Inception-V4}     & 0.819                      & 0.705                      & 0.762                      \\ \hline

\textbf{VGG-16}            & 0.937                      & 0.622                      & 0.782\\ \hline
\textbf{VGGFACE}          & 0.901                      & 0.799                      & 0.852                     \\ \hline
\multicolumn{1}{|c|}{\textbf{AVG}} & \multicolumn{1}{l|}{\textbf{0.886}} & \multicolumn{1}{|c|}{\textbf{0.705}} & \multicolumn{1}{|c|}{\textbf{0.795}} \\ \hline
\end{tabular}%
\label{over_UTK}
}
\end{table}

The goal of this experiment is to evaluate the impact on the fairness of the gender classification algorithms on gender-race groups when the training dataset is skewed towards certain sub-groups. To this aim, the training subset of the UTKFace dataset is used for fine-tuning the gender classification algorithms, which is skewed towards the male population and did not contain Middle Eastern, Latino, and explicit division of Asian. The models are evaluated on the testing part of the FairFace dataset containing gender-balanced seven gender-race groups. As both UTKFace and FairFace datasets are scraped from the web, the cross-dataset impact may not be applicable.

Table~\ref{over_UTK} tabulates the overall performance of the models when fine-tuned on the UTKFace dataset and tested on the FairFace test set. 
It can be seen that the overall performance of all the models dropped. The reason is the under-representation of races and over-representation of the male population in the training set. All the models performed equivalent with an overall accuracy of $0.789$, $0.762$, $0.780$, and $0.850$ for ResNet-50, InceptionNet, VGG-16, and VGGFace, respectively. The overall gap between male and female accuracy rates have increased to $0.181$ from $0.07$ (obtained when a balanced training set was used). VGG-19 obtained almost equal accuracy rates in comparison to VGG-16. 

Table~\ref{subgroup_UTK}, shows the gender classification accuracy across gender-race groups for all the models when trained on the UTKFace dataset. In this case, all the models obtained higher accuracy rates for males over females. This is in contrary to results obtained in Table~\ref{fairFace:genderrace}, where VGG-16 obtained higher accuracy rates for females from all the race groups, except Black females, over males.
For each model, the standard deviation in the accuracy rates across gender and races has increased by at least $0.43$ (Table~\ref{subgroup_UTK1}).
\emph{Middle Eastern} males still obtain higher accuracy rates followed by Indian, Latino, and White males. On an average, Latino females outperformed all other females. This is followed by East Asian and Middle Eastern. The average accuracy for Black females further reduced by $0.143$ and remains the least ($0.606$). 

Table~\ref{FPR_FRR_UTKFACE} shows the false positives and false negatives of the gender classification system when trained on the UTKFace dataset. The highest false positive was obtained for the Black race, which suggests that Black females are most often misclassified as males. This is followed by Southeast Asian females. The highest false negative was obtained by Middle Eastern. 

\emph{These results suggest that a skewed training dataset can further escalate the difference in the accuracy values across gender-race groups}. However, architectural differences and skewed training datasets are not the only reasons for bias in the gender classification system. In fact, for both the experiments (Exp 1 and Exp 2), Black race consistently obtained least performance. Black females consistently obtained lower performance over Black males.  

\begin{table*}[]
\centering
    \caption{Gender classification accuracy of across gender-race groups. Skewed subset of UTKFace dataset was used for fine-tuning the models. The models are evaluated on FairFace balanced test subset.}
\begin{adjustbox}{width=1\textwidth}
\begin{tabular}{|l|c|c|c|c|c|c|c|c|c|c|c|c|c|c|}
\hline
\multicolumn{1}{|c|}{\textbf{Race}}   & \multicolumn{2}{c|}{\textbf{WHITE}}                     & \multicolumn{2}{c|}{\textbf{ME EASTERN}}              & \multicolumn{2}{c|}{\textbf{BLACK}}                     & \multicolumn{2}{c|}{\textbf{EAST ASIAN}}                & \multicolumn{2}{c|}{\textbf{SE ASIAN}}                  & \multicolumn{2}{c|}{\textbf{INDIAN}}                   & \multicolumn{2}{c|}{\textbf{LATINO}}                   \\ \hline
\multicolumn{1}{|c|}{\textbf{Gender}} & \textbf{M}                 & \textbf{F}                 & \textbf{M}                & \textbf{F}                & \textbf{M}                 & \textbf{F}                 & \textbf{M}                 & \textbf{F}                 & \textbf{M}                 & \textbf{F}                 & \textbf{M}                & \textbf{F}                 & \textbf{M}                & \textbf{F}                 \\ \hline
\textbf{ResNet-50}            & 0.911           & 0.683           & 0.919         & 0.735          & 0.864           & 0.583           & 0.858          & 0.731           & 0.848           & 0.707      & 0.890           & 0.672           & 0.907          & 0.737                       \\ \hline

\textbf{Inception-V4}        & 0.826           & 0.720            & 0.867          & 0.755         & 0.826           & 0.589          & 0.797            & 0.693            & 0.789           & 0.696           & 0.829          & 0.678           & 0.798          & 0.805            \\ \hline
\textbf{VGG-16}               & 0.956           & 0.610            & 0.956         & 0.634          & 0.882           & 0.530             & 0.940           & 0.594          & 0.943          & 0.602           & 0.927          & 0.670           & 0.957         & 0.713           \\ \hline
\textbf{VGGFACE}             & 0.929           & 0.784           & 0.942          & 0.823          & 0.821          & 0.724          & 0.905          & 0.816          & 0.906          & 0.775          & 0.890         & 0.826          & 0.914         & 0.842          \\ \hline
\multicolumn{1}{|c|}{\textbf{AVG}}    & \textbf{0.905} & \textbf{0.700} & \textbf{0.921} & \textbf{0.737} & \textbf{0.848 } & \textbf{0.606} & \textbf{0.875} & \textbf{0.708} & \textbf{0.871} & \textbf{0.695} & \textbf{0.884} & \textbf{0.711} & \textbf{0.894} & \textbf{0.774} \\ \hline

\end{tabular}
\end{adjustbox}
\label{subgroup_UTK}
\end{table*}

\begin{table*}[]
\centering
\caption{Statistics of the accuracy values obtained in Table~\ref{subgroup_UTK}.}
\begin{adjustbox}{width=1\columnwidth}
\begin{tabular}{|l|c|c|c|c|c|}
\hline
\multicolumn{1}{|c|}{} & \textbf{MAX} & \textbf{MIN} & \textbf{AVG} & \textbf{STDEV} & \textbf{Diff in AVG} \\ \hline
\textbf{Resnet-50}      & 0.92         & 0.583        & 0.79         & 0.108          & 1.35                 \\ \hline
\textbf{Inception-V4}  & 0.8672       & 0.59         & 0.762        & 0.077          & 0.797                \\ \hline
\textbf{VGG-16}         & 0.957        & 0.53         & 0.78         & 0.169          & 2.2089               \\ \hline
\textbf{VGGFACE}       & 0.9422       & 0.724        & 0.85         & 0.0655         & 0.716                \\ \hline
\end{tabular}%
\end{adjustbox}
\label{subgroup_UTK1}
\end{table*}

\begin{table*}[]
\centering
\caption{False positives and negatives measured for accuracy values in Table~\ref{subgroup_UTK}. False positives are females misclassified as males and false negatives are males misclassified as females.}
\begin{adjustbox}{width=1\textwidth}
\begin{tabular}{|l|c|c|c|c|c|c|c|c|c|c|c|c|c|c|}
\hline
\multicolumn{1}{|r|}{\textbf{Race}} & \multicolumn{2}{c|}{\textbf{WHITE}}                                                               & \multicolumn{2}{c|}{\textbf{ME EASTERN}}                                                          & \multicolumn{2}{c|}{\textbf{BLACK}}                                                               & \multicolumn{2}{c|}{\textbf{EAST ASIAN}}                                                          & \multicolumn{2}{c|}{\textbf{SE ASIAN}}                                                            & \multicolumn{2}{c|}{\textbf{INDIAN}}                                                             & \multicolumn{2}{c|}{\textbf{LATINO}}                                                              \\ \hline
\textbf{}                           & \textbf{FP}                                    & \textbf{FN}                                    & \textbf{FP}                                    & \textbf{FN}                                    & \textbf{FP}                                    & \textbf{FN}                                    & \textbf{FP}                                    & \textbf{FN}                                    & \textbf{FP}                                    & \textbf{FN}                                    & \textbf{FP}                                    & \textbf{FN}                                   & \textbf{FP}                                    & \textbf{FNR}                                    \\ \hline
\textbf{ResNet-50}                   & 0.2298                               & 0.132                                & 0.123                                & 0.184                                & 0.314                                & 0.198                                & 0.238                               & 0.163                               & 0.242                               & 0.189                                & 0.272                               & 0.139                            & 0.233                                & 0.108                                 \\ \hline
\textbf{Inception-V4}               & 0.225                              & 0.223                                 & 0.121                               & 0.265                             & 0.320                               & 0.238                               & 0.277                                & 0.228                             & 0.263                               & 0.247                              & 0.283                                & 0.199                             & 0.204                                & 0.193                              \\ \hline
\textbf{VGG-16}                      & 0.39                                 & 0.0436                                & 0.366                                & 0.634                                & 0.472                                 & 0.118                                & 0.406                                & 0.062                                & 0.398                                & 0.057                               & 0.330                                & 0.073                               & 0.287                                & 0.043                              \\ \hline
\textbf{VGGFACE}                    & 0.216                               & 0.071                               & 0.177                               & 0.058                                & 0.2761                               & 0.179                                 & 0.184                               & 0.095                              & 0.225                               & 0.094                              & 0.174                                & 0.110                              & 0.158                               & 0.085                             \\ \hline
\multicolumn{1}{|c|}{\textbf{AVG}}           & \multicolumn{1}{l|}{\textbf{0.265}} & \multicolumn{1}{l|}{\textbf{0.117}} & \multicolumn{1}{l|}{\textbf{0.197}} & \multicolumn{1}{l|}{\textbf{0.285}} & \multicolumn{1}{l|}{\textbf{0.346}} & \multicolumn{1}{l|}{\textbf{0.183}} & \multicolumn{1}{l|}{\textbf{0.276}} & \multicolumn{1}{l|}{\textbf{0.137}} & \multicolumn{1}{l|}{\textbf{0.282}} & \multicolumn{1}{l|}{\textbf{0.147}} & \multicolumn{1}{l|}{\textbf{0.265}} & \multicolumn{1}{l|}{\textbf{0.131}} & \multicolumn{1}{l|}{\textbf{0.221}} & \multicolumn{1}{l|}{\textbf{0.107}} \\ \hline
\end{tabular}%
\end{adjustbox}
\label{FPR_FRR_UTKFACE}
\end{table*}

\begin{figure}
    \centering
    \includegraphics[width=0.7\columnwidth]{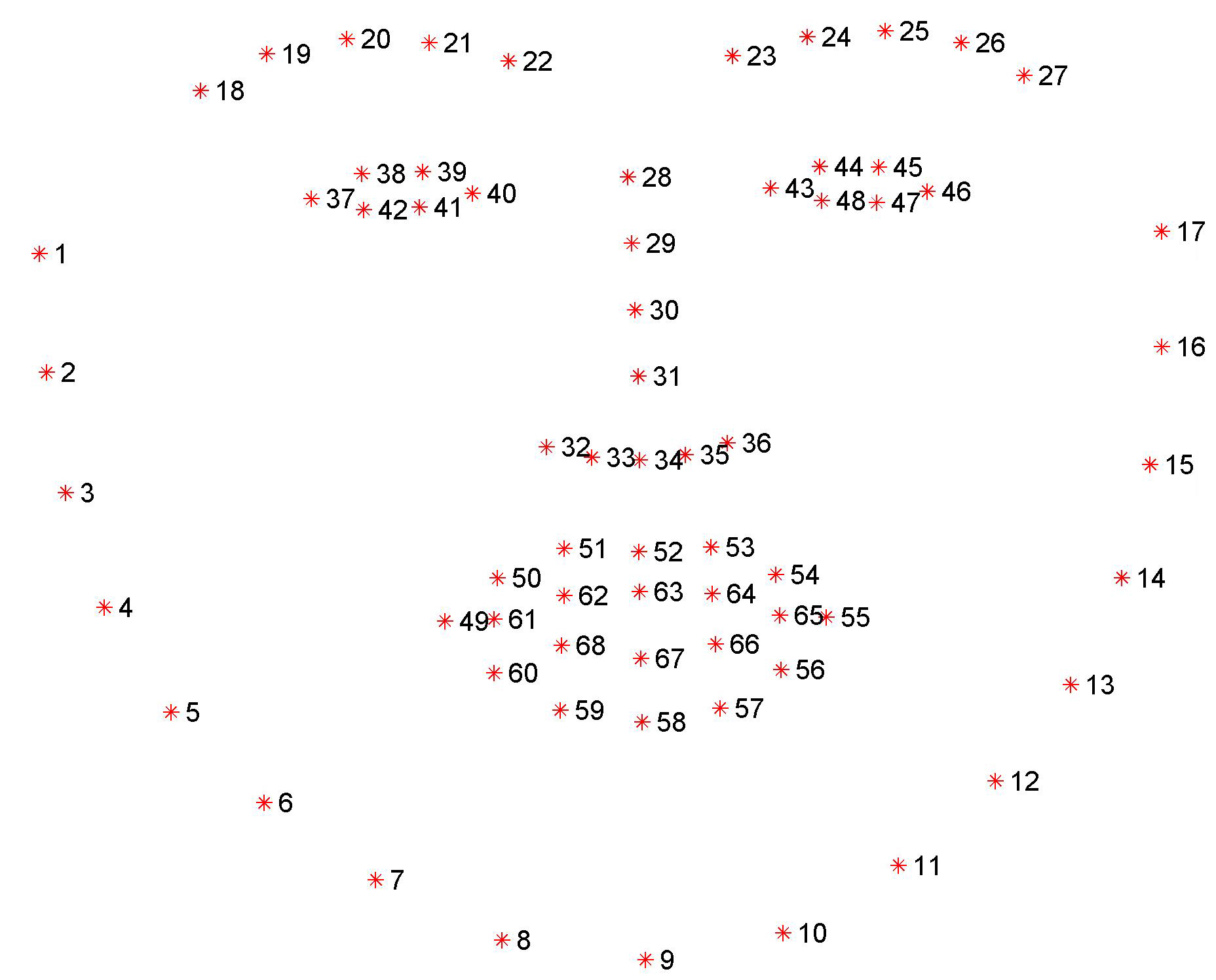}
    \caption{Visualizing the 68 facial landmark coordinates from Dlib landmark detector~\cite{dlib}.}
    \label{dlib_land}
\end{figure}

\noindent \textbf{Exp \#3: Facial Morphological Differences:}
To further understand the reason for the consistent low accuracy of Black race vs. others and Black females (Exp \#1 and Exp \#2), in particular, we studied the difference in the facial morphology between the Black race and others.
To this aim, we randomly selected $500$ male and $500$ female face images for each of the seven-race groups from the FairFace dataset using python script and extracted $68$ facial landmarks using Dlib~\cite{dlib} library. Figure~\ref{dlib_land} shows the indexes of the $68$ landmark coordinates visualized on the image. 

The $68$ landmark locations for each face images were appended together into a one-dimensional feature vector of $128$ dimensions. The feature vectors are clustered using K-means clustering for understanding differences in facial morphology. 

\begin{figure} 
    \centering
    \includegraphics[width=0.7\columnwidth]{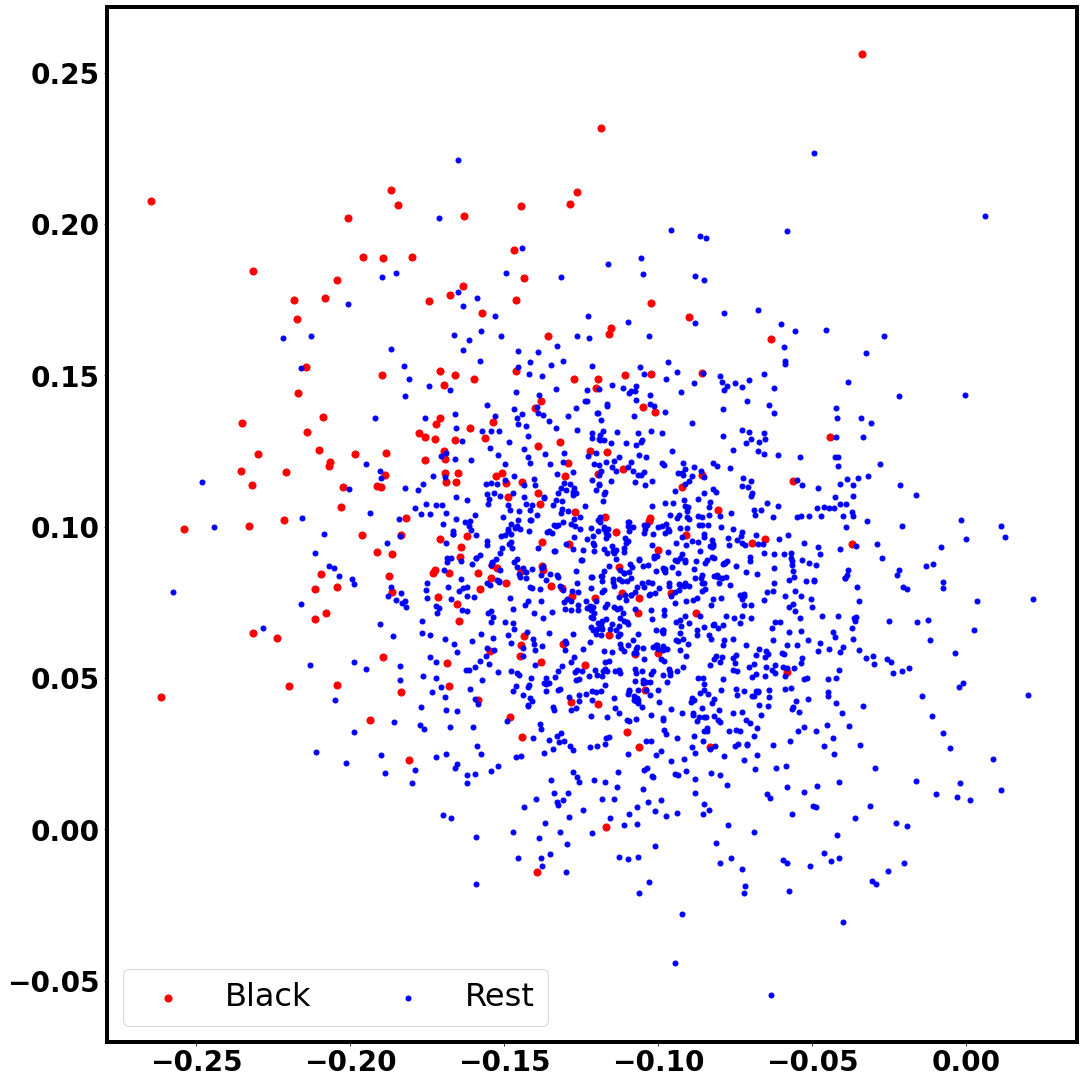} 
    \caption{Clustering of facial landmarks of all races into two groups. $92\%$ of the Black males and females were clustered together in one of the groups. $62.8\%$ of the facial landmarks belonging to other races were classified together into second group.}
    \label{BvsAR}
\end{figure}

\begin{figure}
    \centering
    \includegraphics[width=0.7\columnwidth]{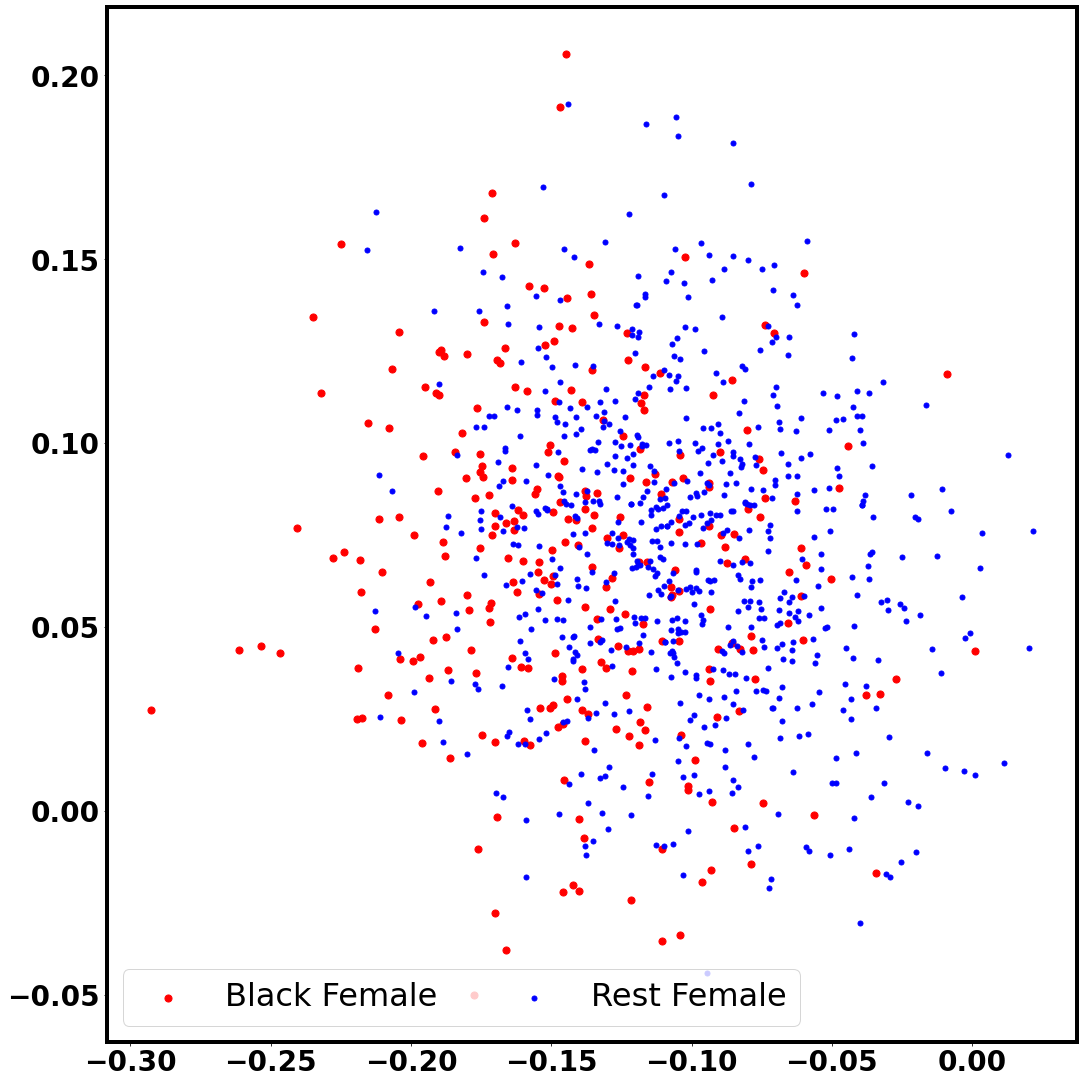}
    \caption{Clustering of facial landmarks of all females into two groups. $96.8\%$ Black females were grouped together in a different cluster. $63.16\%$ of the facial landmarks belonging to other females were classified together into second group.}
    \label{BFRF}
\end{figure}

\begin{figure}
    \centering
    \includegraphics[width=0.7\columnwidth]{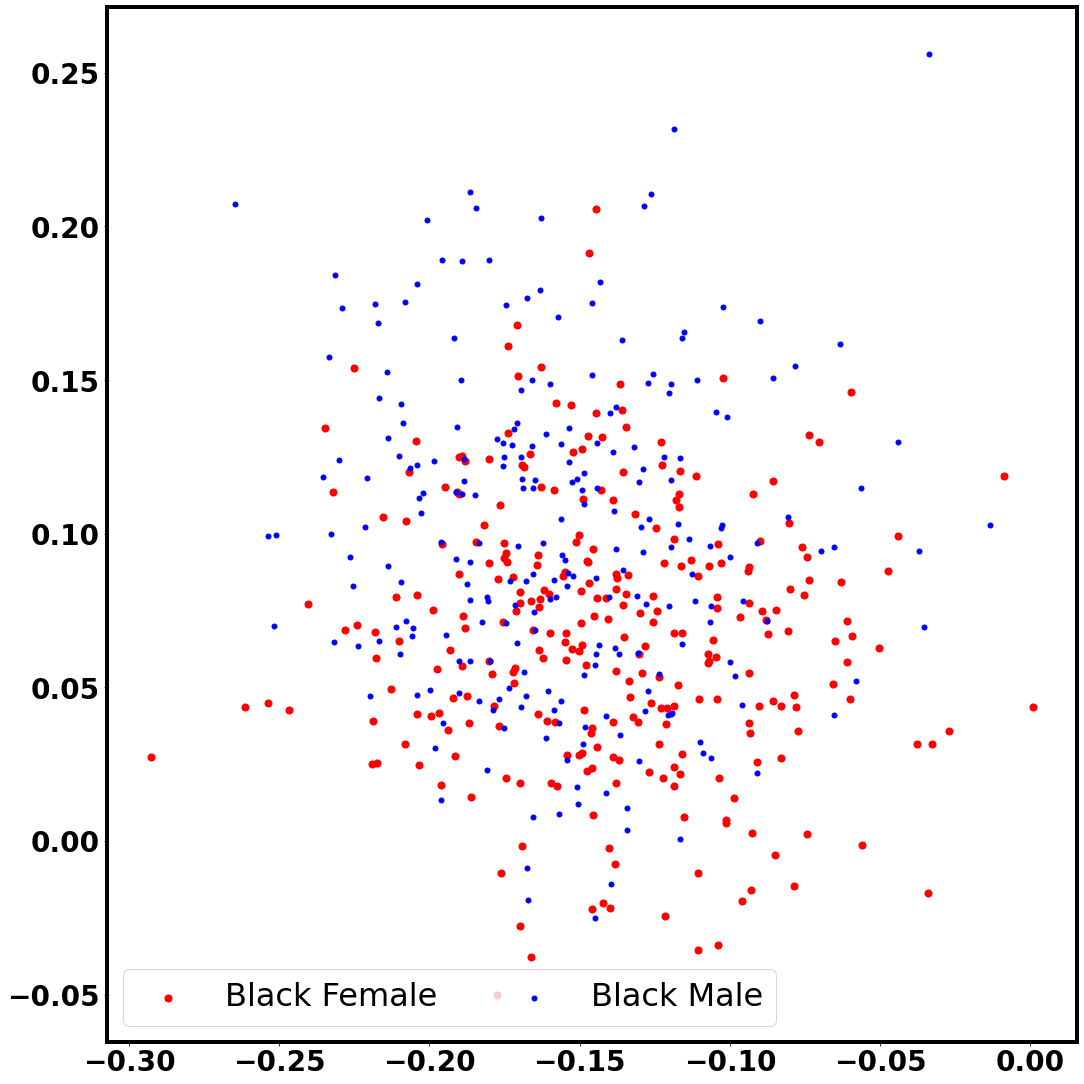}
    \caption{Clustering of facial landmarks of Black females and Black males. Only $11.7\%$ and $35.2\%$ of the landmarks belonging to Black females and males, respectively, were grouped together into another cluster which suggest high morphological similarity between Black males and females.}
    \label{BFBM}
\end{figure}

Figure~\ref{BvsAR} shows the plot obtained on clustering the facial landmarks from all the races into two clusters. Among all, $92\%$ of the Black males and females were clustered together. $62.8\%$ of the facial landmarks belonging to other races were clustered together into the second group. Figure~\ref{BFRF} shows the clustering of facial landmarks from females of all the races into two clusters. $96.8\%$ of the Black females were grouped together in a single cluster. $63.16\%$ of the facial landmarks belonging to females from other race were clustered together into the second group. Figure~\ref{BFBM} shows the plot obtained on clustering the facial landmarks of Black males and females into two clusters. Among all the samples, only $11.7\%$ and $35.2\%$ of the landmarks belonging to Black females and males, respectively, were grouped together into a single cluster, which suggests a high facial morphological similarity between Black males and females.

The above plots suggest that significant facial morphological differences are the result of consistent low accuracy rates of the Black race. \emph{These results also suggest that high morphological similarity between Black males and females are the potential cause of the least accuracy rates for Black females}. 

\section{Conclusion and future work}
In this paper, we investigated the source of bias of the gender classification algorithms across gender-race groups. Experimental investigations suggested that algorithms with architectural differences may vary in performance even when trained on race and gender-balanced set. Therefore,  the  bias  of  the  gender classification system is not due to a particular algorithm. For all the experiments conducted, Black Race and Black females in specific obtained least accuracy rates. Middle Eastern males and Latino females obtained highest accuracy rates, also observed in~\cite{krkkinen2019fairface}. The reason could be skin-tone reflectance property in varying illumination combined with facial morphology. The skewed training set can further increase the inequality in the accuracy rates. Further, the analysis suggested that facial morphological differences between Black females and the rest females, and high similarity with the Black males could be the potential cause of their high error rates. 

As a part of future work, statistical validation of the results will be conducted on other datasets. The impact of other covariates such as pose, illumination, and make-up on unequal accuracy rates will be studied. 
Experiments on facial morphological differences will be extended using deep learning-based landmark localization methods~\cite{7553523} on all gender-race groups. The reason for a specific gender-race group outperforming others will be investigated based on skin-tone reflectance property, facial morphology, and the impact of other covariates.


\bibliographystyle{IEEEtran}
\bibliography{biblio}

\end{document}